\documentclass[preprint]{elsarticle}
\usepackage{amssymb}
\usepackage{algorithm2e}
\usepackage{amsthm}
\usepackage{amsmath}

\usepackage[T1]{fontenc} 
\usepackage{booktabs}
\usepackage{hyperref}
\usepackage{caption}
\usepackage{xcolor}
\usepackage{float}
\usepackage{amsfonts}

\makeatletter
\def\ps@pprintTitle{%
  \let\@oddhead\@empty
  \let\@evenhead\@empty
  \let\@oddfoot\@empty
  \let\@evenfoot\@oddfoot
}
\makeatother

\biboptions{numbers,sort&compress}
\DeclareMathAlphabet{\mathbbold}{U}{bbold}{m}{n}

\AtBeginDocument{}
\DeclareMathOperator*{\argmin}{argmin} 

\begin{document}
\captionsetup[figure]{labelfont={bf},labelformat={default},labelsep=period,name={Fig.}}

\newcommand{\JB}[1]{{\color{red} #1}}
\newcommand{\CB}[1]{{\color{blue} [CB: #1]}}
\newcommand{\PT}[1]{{\color{cyan} #1}}

\begin{frontmatter}

\title{Comparative study of machine learning and statistical methods for automatic identification and quantification in $\gamma$-ray spectrometry}

\author[inst1]{Dinh Triem Phan\corref{cor1}}
\cortext[cor1]{Corresponding author}

\ead{dinh-triem.phan@cea.fr}
\affiliation[inst1]{organization={Université Paris-Saclay, CEA, List, Laboratoire national Henri Becquerel (LNE-LNHB)},
            city={Palaiseau},
            postcode={91120}, 
            country={France}}

\author[inst2]{Jérôme Bobin}
\author[inst1]{Cheick Thiam}
\author[inst1]{Christophe Bobin}
\affiliation[inst2]{organization={IRFU, CEA, Universite Paris-Saclay},
            city={Gif-sur-Yvette},
            postcode={91191}, 
            country={France}}

\begin{abstract}
 
During the last decade, a large number of different numerical methods have been proposed to tackle the automatic identification and quantification in $\gamma$-ray spectrometry. However, the lack of common benchmarks, including datasets, code and comparison metrics, makes their evaluation and comparison hard. In that context, we propose an open-source benchmark that comprises simulated datasets of various $\gamma$-spectrometry settings, codes of different analysis approaches and evaluation metrics. This allows us to compare the state-of-the-art end-to-end machine learning with a statistical unmixing approach using the full spectrum. Three scenarios have been investigated: (1) spectral signatures are assumed to be known; (2) spectral signatures are deformed due to physical phenomena such as Compton scattering and attenuation; and (3) spectral signatures are shifted ({\it e.g.}, due to temperature variation). A large dataset of $2.10^5$ simulated spectra containing nine radionuclides with an experimental natural background is used for each scenario with multiple radionuclides present in the spectrum. Regarding identification performance, the statistical approach consistently outperforms the machine learning approaches across all three scenarios for all comparison metrics. However, the performance of the statistical approach can be significantly impacted when spectral signatures are not modeled correctly. Consequently, the full-spectrum statistical approach is most effective with known or well-modeled spectral signatures, while end-to-end machine learning is a good alternative when measurement conditions are uncertain for radionuclide identification. Concerning the quantification task, the statistical approach provides accurate estimates of radionuclide counting, while the machine learning methods deliver less satisfactory results.  
\end{abstract}

\end{frontmatter}


\section{Introduction}

Gamma-ray spectrometry is a widely used technique for identifying and quantifying $\gamma$-emitting radionuclides, with various applications in nuclear physics including nuclear security, environmental monitoring and radiological characterization for decommissioning of nuclear facilities. Specific applications, such as in situ environmental analysis following a nuclear accident or the detection of illicit nuclear material trafficking, require rapid and reliable radionuclide identification under challenging conditions, including short measurement durations ({\it i.e.,} low statistics). Meeting these demands necessitates methods that are accessible to non-experts and capable of making robust decisions. Traditional methods \cite{gilmore2008practical}, such as peak-based regression assuming Gaussian noise statistics, rely heavily on expert intervention and struggle to address these challenges, particularly in scenarios involving complex mixtures of radionuclides or low count rates.
 
Recent advances in machine learning (ML) have been applied to $\gamma$-ray spectrometry, offering automated, end-to-end solutions to overcome existing challenges. Techniques such as convolutional neural networks (CNN) and multi-layer perceptrons (MLP) \cite{liang2019rapid,daniel2020automatic,kim2019multi,koo2021development, turner2021convolutional, kim2025deep,kamuda2020comparison,kamuda2017automated, kamuda2019automated} utilize large training datasets of different radionuclide mixtures to automate the identification and quantification process. These methods take $\gamma$-spectra as input, the presence of radionuclides as output for identification and their mixing weights as output for quantification. ML approaches have been effectively applied across various detectors and applications, yielding robust results in radionuclide identification. 

Besides this approach, the statistical method based on full-spectrum analysis with Poisson likelihood has also shown solid performance. A notable example is the full-spectrum unmixing technique, which models an observed $\gamma$-spectrum $y \in \mathbb{R}^{M}$ where $M$ is the number of channels as a Poisson distribution of $Xa$:
\begin{equation}
y \sim \mathcal{P}(Xa)
\label{eq:poisson}
\end{equation}
Here, $X \in \mathbb{R}^{M\times N} $ is a matrix of the normalized spectral signature of all $N$ radionuclides, including the natural background (Bkg). Each column of $X$ represents the detector’s characteristic response to $\gamma$-photon emissions from a specific radionuclide. The vector $a \in \mathbb{R} ^{N}$ contains the corresponding counting for each radionuclide. A regularized Maximum Likelihood Estimation (MLE) is then employed to estimate the counting of all radionuclides. This method has been successfully applied to NaI(Tl) and HPGe detectors and demonstrated to give better performance in activity estimation and decision thresholds than traditional methods \cite{xu2020sparse,malfrait2023spectral, malfrait2023online, mano2024algorithm}. For instance, experimental results obtained on real aerosol filter samples revealed that this approach enables rapid detection of $^{137}$Cs at extremely low activity levels (mBq) around 1.5 days after sampling, while traditional methods typically require around eight days \cite{malfrait2023spectral, malfrait2023online}.

Most existing studies lack publicly available code and datasets, making it difficult to reproduce results and perform fair comparisons between methods. To address this issue, we introduce an open-source benchmark that includes the full implementation of ML and spectral unmixing methods, as well as a simulated dataset and standardized evaluation metrics. This benchmark establishes a framework for fair and consistent comparison while allowing easy integration of new methods and flexible adaptation to various datasets and use cases.

The benchmark compares the ML and statistical unmixing approaches in three scenarios: (1) spectral signatures are assumed to be known, corresponding to well-defined measurement conditions; (2) spectral signatures are deformed due to physical phenomena such as Compton scattering and attenuation; and (3) gain shift ({\it e.g.}, due to temperature variations). The evaluation uses a large dataset of 200000 simulated spectra containing nine radionuclides with an experimental natural background (Bkg). These simulated spectra are generated by varying the number of radionuclides present and the counting associated with each radionuclide. Both methods are evaluated based on their performance in identifying and quantifying $\gamma$-emitting radionuclides. To ensure fairness in comparison, the CNN architecture and hyperparameters are optimized, and classification thresholds are calibrated to achieve a false positive rate close to the predefined value.

All materials for this work are available on the \href{https://github.com/triem1998/BenchmarkGamma/}{GammaBench} GitHub repository. This includes the Geant4-simulated spectral signatures for all radionuclides across the three scenarios. The repository also provides the full codebase developed in Python using the open-source PyTorch library \cite{NEURIPS2019_9015}, covering data generation, ML architectures, hyperparameter optimization, pre-trained models and spectral unmixing algorithms.

The article is organized as follows:
\begin{itemize}
\item {Section 2 provides a description of the dataset and the evaluation metrics for identification and quantification.}
\item {Section 3 presents the end-to-end ML and statistical unmixing full-spectrum approaches }
\item {Section 4 shows the results of these two approaches for three scenarios for identification and quantification.}

\end{itemize}

\section{Datasets and evaluation metrics}

\subsection{Datasets }\label{sec:data}

Automatic identification and quantification in $\gamma$-ray spectrometry can be applied using different types of detectors, including CdTe, LaBr3, NaI(Tl) and HPGe detectors. In that context, NaI(Tl) detectors are widely used due to their low cost and high detection efficiency allowing short measurement time. As a result, this study utilizes a 3"$\times$3" NaI(Tl) detector to compare ML and statistical unmixing methods. While the analysis focuses on this specific detector, the methodologies presented are readily adaptable to other detector types with different energy resolutions.

In general, training an ML model requires a large and representative dataset that captures all key parameters characterizing the data. In the context of radionuclide identification and quantification, this means including a wide range of $\gamma$-ray spectra that account for both the diversity of radionuclides and their varying contributions. Obtaining such a dataset experimentally can be difficult and time-consuming due to the complexity of possible spectral combinations. To address this, a common approach in the literature is to simulate spectral signatures for various radionuclides using Monte Carlo codes such as Geant4 \cite{agostinelli2003geant4} or MCNP \cite{forster2006mcnp}. Synthetic spectra are then generated by following the Poisson distribution of a linear combination of normalized spectral signatures. The synthetic dataset is constructed by randomly selecting the number of radionuclides, the total counting (counting of all radionuclides) and the mixing weights representing each radionuclide's contribution.

Three datasets were created to address different scenarios: (1) ideal conditions with known spectral signatures, (2) deformed spectral signatures due to physical phenomena and (3) gain shift.

\begin{figure}[H]
    \centering
      \includegraphics[scale=0.18]{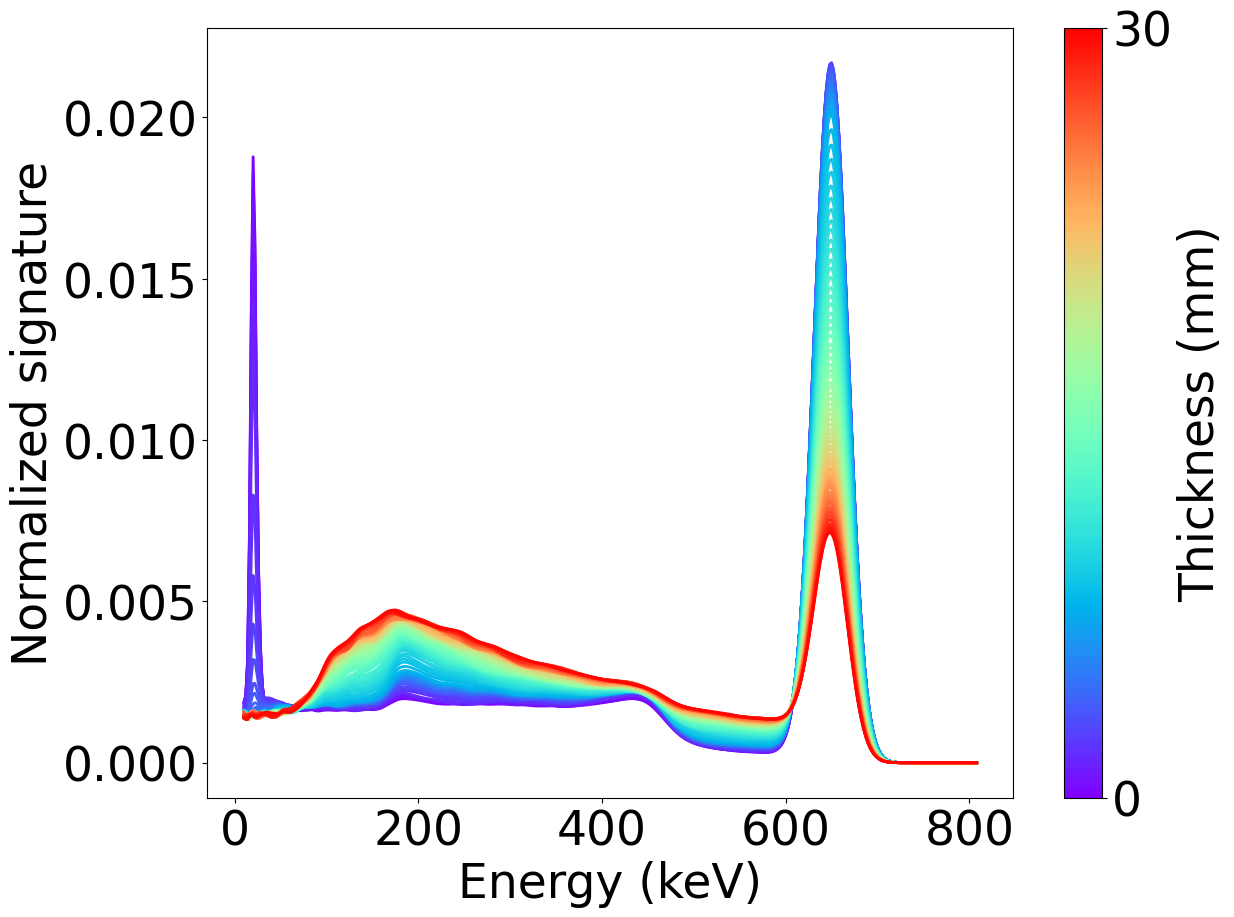}
    \includegraphics[scale=0.18]{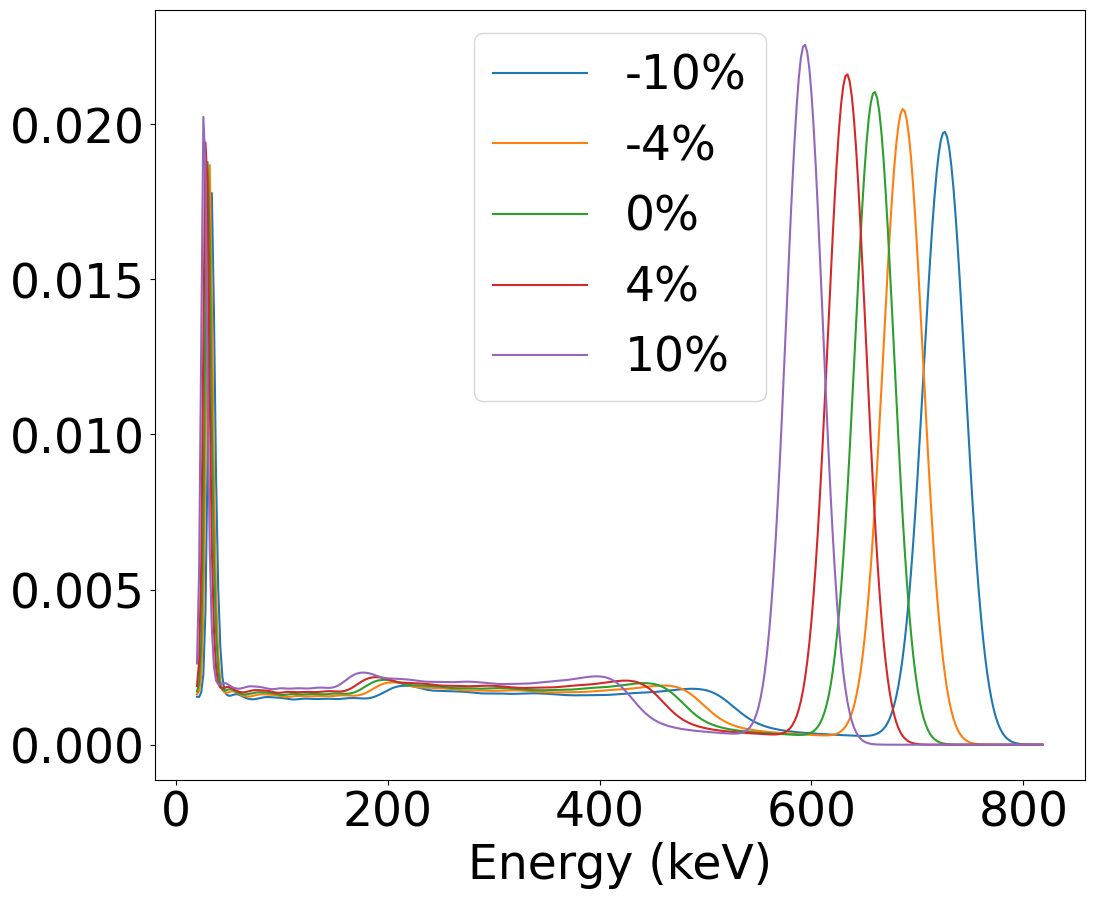}
    \caption{The left panel (Scenario 2) shows spectral signatures of $^{137}$Cs as a function of steel thickness, with deformation caused by $\gamma$-photon interactions with the surrounding sphere. The right panel (Scenario 3) displays spectral signatures of $^{137}$Cs as a function of gain shift.}
      \label{figure:spec_cs137}
\end{figure}

\paragraph{\bf Scenario 1: Known spectral signatures}
This scenario represents idealized conditions where measurement conditions are well-defined and spectral signatures are assumed to be known. The library of radionuclides has nine radionuclides covering a large range of energies between 20 keV and 2 MeV: $^{57}$Co, $^{60}$Co,  $^{99m}$Tc, $^{123}$I, $^{131}$I, $^{133}$Ba,  $^{137}$Cs, $^{152}$Eu and $^{241}$Am. Most of these radionuclides are classical $\gamma$-emitters for handheld instruments or spectroscopy-based portal monitors (e.g., ANSI N42.38-2015 \cite{7394937}) and widely used in the literature. Spectral signatures were generated from Geant4 \cite{agostinelli2003geant4}, a Monte Carlo toolkit for simulating particle interactions with matter. A point source was located 20 cm from the 3"$\times$3" NaI(Tl) detector. The energy resolution of the NaI(Tl) detector was considered (6.5$\%$ at 662 keV). Each spectral signature consists of 1024 channels, with a binning of 2 keV per channel and a low-energy cut-off of 20 keV. 
\paragraph{\bf Scenario 2: Deformed spectral signatures}
This scenario illustrates the deformation of spectral signatures when a source is enclosed within a steel sphere of varying thickness. Although conceptually simple, it effectively captures the impact of physical phenomena-such as attenuation and Compton scattering-resulting from the interaction of $\gamma$-photon with the surrounding steel. The dataset is based on simulations presented in \cite{phan2024hybrid}, where a point source was placed inside a steel sphere with thickness varying from 0.001 mm to 30 mm. In this case, 96 spectral signatures of each radionuclide were simulated by Geant4 for the NaI(Tl) detector. Due to the complexity of spectral variability, the library was reduced from 9 radionuclides in the previous scenario to 8 by excluding $^{131}$I from the originally defined library. The evolution of the spectral signatures of $^{137}$Cs as a function of the sphere's thickness is shown in \autoref{figure:spec_cs137}.

\paragraph{\bf Scenario 3: Gain shift}

Another factor contributing to spectral variability is gain shift, which can be caused, for example, by temperature variation in scintillation detectors. This gain shift phenomenon has also been studied in recent literature for its impact on automatic radionuclide identification and quantification \cite{kamuda2019automated,liang2019rapid,turner2021convolutional,kim2025deep}. For this scenario, the same radionuclides and spectral signatures as those used in the known-signatures case were applied. For each radionuclide, Geant4 provides a list of deposited energies, and the spectral signature is represented as a histogram of these energies across a specific energy range. The shifted spectral signatures are described by the relation: $e_s= e_r \times (1-\alpha)$ where $e_s$ is the shifted energy, $e_r$ is the energy of the reference spectral signature, and $\alpha$ is the shift factor. The shifted spectral signatures are then constructed based on these shifted energies. An example of spectral signatures of $^{137}$Cs with different values of the shift factor $\alpha$ is shown in \autoref{figure:spec_cs137}.

All three scenarios discussed above were used to create synthetic datasets. For each scenario, a dataset consisting of 200000 simulated spectra was generated through the linear combination of spectral signatures (including Bkg) and Poisson noise. In this work, Bkg was supposed to be known. The total counting was sampled from a log-uniform distribution ranging from 200 to 100000: $log(\sum a) \sim U(2+log(2),5)$. The number of present radionuclides in the simulated spectra was randomly chosen between 0 and 4. Bkg is always present in the spectrum and contributes at least 10$\%$. The mixing weights $z=a/ \sum(a)$ of the radionuclides were selected randomly, with the constraint that their sum equals one. The minimum counting for $^{60}$Co, $^{137}$Cs and $^{152}$Eu is 100, while for other radionuclides, the minimum is 50. The dataset was then divided into training, validation, and test sets in a 64:16:20 ratio. For the spectral deformation scenario, the thicknesses of the steel spheres were randomly selected from 96 values ranging from 0.001 mm to 30 mm. In the case of spectral shifts, the shift factor was uniformly varied between -10$\%$ and 10$\%$, accounting for a wide range of potential shifts.

\subsection{Evaluation metrics}
\subsubsection{Identification}
For the identification task, let $(y_1,l_1),..,(y_K,l_K) $ represent the test set where $y_i$ is the $i$-th $\gamma$-spectrum in test set and  $l_i$ is its associated label. The label $l_i=(l_{i,2},..,l_{i,N})$ is defined such that $l_{i,j}=1$  if the radionuclide $j$ is present in the spectrum $y_i$ and $l_{i,j}=0$ otherwise. The index $j$ starts from 2 since Bkg is always present in the simulated spectrum. Concerning the regression task, $(y_i,z_i) $ is the $i$-the $\gamma$-spectrum and its corresponding mixing weight. The mixing weight $z_i=(z_{i,1},..,z_{i,N})$ is defined as the radionuclide counting divided by the total counting $z_{i,j}=a_{i,j}/\sum_{j=1}^N (a_{i,j})$. 

Since radionuclide identification is a multi-label classification problem (multiple radionuclides (labels) can be present in a single $\gamma$-spectrum), the following metrics are used for performance evaluation:
\begin{itemize}
    \item False positive rate (FPR) for all radionuclides: the proportion of cases where at least one radionuclide is incorrectly predicted to be present while it is absent.
    \item False negative rate (FNR) for all radionuclides: the proportion of cases where at least one radionuclide present in the radioactive source is not identified.
    \item Perfect prediction rate (PPR): the proportion of spectra in the dataset with no false positives or negatives. It is defined as: $PPR=(\sum _{i=1}^K  \mathbbold{1} _{\hat l_i=l_i})/K$ where $\hat l$ is the label predicted by the model.
    \item False prediction rate (FPrR) for all radionuclides: the proportion of spectra with any false predictions, including false positives, false negatives, or both. This is related to PPR as: $FPrR=1-PPR$
    \item Accuracy: for each radionuclide, the accuracy is defined as the ratio of correct predictions to the total number of samples. The accuracy for the multi-label problem is calculated as the average accuracy across all radionuclides: $acc=1/(N-1) \times \sum _{j=2}^{N} (\sum _{i=1}^K \mathbbold{1} _{\hat l_{i,j}=l_{i,j}})/K  $
    \item Recall (probability of detection) for each radionuclide: proportion of present radionuclides that are correctly identified:  $Re_j= \sum _{i=1}^K 1/K \times \mathbbold{1} _{\hat l_{i,j}=1| l_{i,j}=1} $    
\end{itemize}
While additional metrics like precision and F1-score can also be used, they can be derived from the above-mentioned metrics and are not explicitly included in this comparison.

Let us notice that for multi-label classification, the interpretation of accuracy differs from that of binary or multi-class classification. Each radionuclide is associated with a confusion matrix, and accuracy is calculated as the average of the binary classification accuracy for all radionuclides. Radionuclides are present much less often than they are absent, with a maximum of four out of nine appearing at the same time. In this dataset, each radionuclide is present in only around 20$\%$ of the cases, creating an imbalance among the labels. As a result, a high accuracy score can be misleading - for example, a naive model predicting the absence of all radionuclides would still achieve 80$\%$ accuracy.

\subsubsection{Quantification}

For quantification tasks, standard regression metrics such as relative error and mean squared error (MSE) are used to measure the difference between the estimated and true values of mixing weights or counting.
\begin{itemize}
    \item MSE on mixing weights: $\sum _{i=1}^K 1/K \times \sum _{j=1}^{N} 1/N \times (\hat z_{i,j} -z_{i,j})^2$ where $\hat z$ is the estimated mixing weight.
    \item Relative absolute error on counting for present radionuclides: $\sum _{i=1}^K 1/K \times  \sum _{j\in S(i)}  |\hat a_{i,j} -a_{i,j}|/a_{i,j}/dim(S(i))$ where  $S(i)$ is the set of present radionuclide for the ith spectrum.
    \item Relative error for individual radionuclide: The relative error is further analyzed for each radionuclide as a function of radionuclide counting.
\end{itemize}

\section{Automatic identification and quantification methods in $\gamma$-ray spectrometry}

\subsection{Machine learning method}

\subsubsection{State of the art}

Radionuclide identification can be considered as a multi-label classification problem in ML where multiple radionuclides (labels) can be present in a single $\gamma$-spectrum. This differs from multi-class classification, which assumes that each spectrum contains only one radionuclide. In the state of the art of ML, several approaches have been proposed to address this multi-label problem \cite{tsoumakas2008multi,read2014deep}.

One widely used approach is problem transformation, which converts the multi-label classification problem into simpler tasks. For instance, the binary relevance method treats each label independently, creating a binary classification problem for each radionuclide (in our case, each model predicts whether a specific radionuclide is present or absent). While straightforward, this method requires training multiple models and does not capture label dependencies ({\it e.g.,} correlation between radionuclides). Some works in $\gamma$-ray spectrometry used this technique are \cite{liang2019rapid,daniel2020automatic}. Another problem transformation technique is Label Powerset, which converts the task into a single-label classification problem in which each unique combination of labels is treated as a single class. Although effective for small numbers of radionuclides, it becomes impractical with a large label set ({\it e.g.,} $2^N$ possible combinations for $N$ labels (radionuclides)). This method has been employed in $\gamma$-ray spectrometry with four radionuclides in \cite{kim2019multi}. Less common transformation methods include Classifier Chains and Pairwise Transformation, which have yet to see significant application in $\gamma$-ray spectrometry.

The second major approach is algorithm adaptation, which modifies existing ML algorithms to handle multi-label classification directly. This approach has been applied to traditional ML methods such as K-nearest neighbors, logistic regression, decision trees and support vector machines. For deep learning models, a common method is to use an average of binary cross-entropy (BCE) over all radionuclides as the loss function with sigmoid activation for the output layer. This configuration allows the network to simultaneously account for all radionuclides and produce outputs between 0 and 1 for each radionuclide, which can then be used to identify radionuclides. This method has been widely employed in recent $\gamma$-ray spectrometry studies \cite{koo2021development, turner2021convolutional, kim2025deep}. Additionally, in multi-label classification, recent studies have explored adaptations of loss functions in Deep Learning such as Focal Loss \cite{lin2017focal} and Asymmetric Loss \cite{ridnik2021asymmetric}. However, these new techniques have not yet been applied to $\gamma$-ray spectrometry.

While radionuclide identification benefits from the detection of peaks (similar to regions of interest), radionuclide quantification is challenging. Few studies in the literature have focused on $\gamma$-ray spectrometry quantification \cite{kamuda2019automated, kamuda2020comparison, kim2025deep}, which involves estimating mixing weights. This task is treated as a regression problem in ML, where common loss functions are mean square error (MSE) or BCE with softmax activation, ensuring that the sum of the mixing weights equals one. Quantification results can also be used for identification by applying appropriate thresholds \cite{kamuda2017automated, kamuda2019automated}.

Concerning ML algorithms used in the $\gamma$-ray spectrometry literature, traditional ML methods have been employed previously. Recently, deep learning approaches such as multilayer perception (MLP) \cite{kamuda2017automated,kamuda2019automated, kim2019multi} and convolutional neural network (CNN) \cite{kamuda2020comparison, daniel2020automatic,liang2019rapid, koo2021development, turner2021convolutional, kim2025deep} have become increasingly popular. Among these, the CNN architecture has been demonstrated to provide better performance with high accuracy in radionuclide identification in recent studies and is then used in this work.

\subsubsection{Machine learning methods in the benchmark}
Two commonly used approaches for automatic identification in $\gamma$-ray spectrometry, binary relevance and multi-label classification with the BCE loss function, were evaluated in this work. For simplicity, multi-label classification applying the BCE loss function is referred to as “multi-label”, and the binary relevance method is referred to as “mono-label” in the remainder of this study.

\begin{figure}[H]
    \centering
      \includegraphics[scale=0.3]{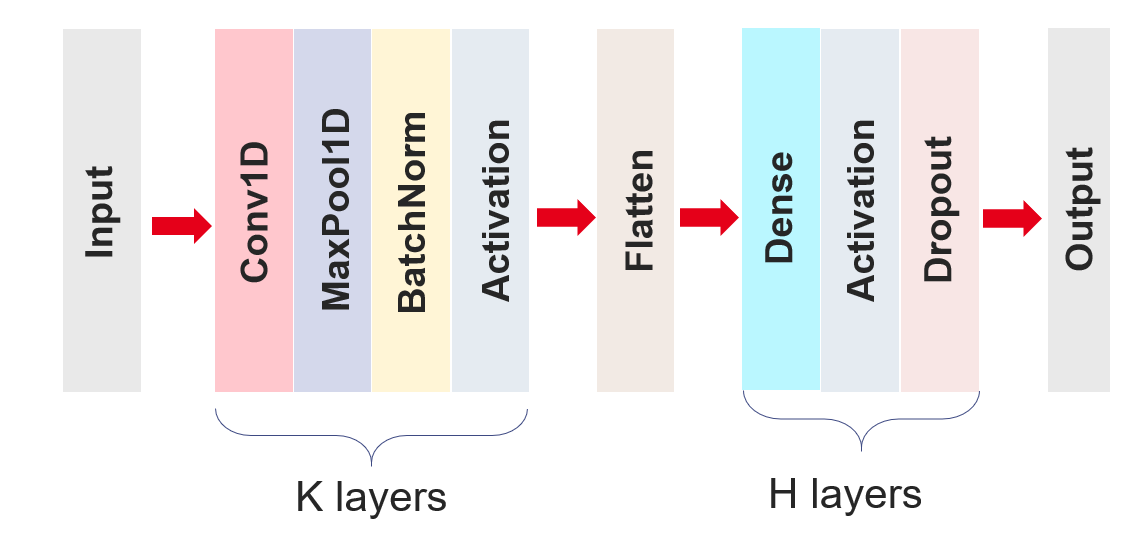}
    \caption{The CNN architecture consists of several convolutional layers followed by fully connected layers.}
      \label{figure:archi}
\end{figure}

The architecture employed is CNN, which is composed of several convolutional layers followed by fully connected layers (FCL). This well-established architecture is commonly applied in ML for classification and has been widely adopted in $\gamma$-ray spectrometry research \cite{daniel2020automatic,koo2021development,turner2021convolutional,kim2025deep,kamuda2020comparison}. The CNN structure is illustrated in \autoref{figure:archi}. Each convolutional layer is designed with the following components:
\begin{itemize}
    \item A 1D convolutional layer to extract useful features such as peaks and attenuation patterns.
    \item A max-pooling layer to reduce spatial dimensions.
    \item Batch normalization to help train faster and in a more stable manner.
    \item ReLU as the activation function.
\end{itemize}
The output of the convolutional layers is flattened and passed through several FCL (dense) layers. Dropout is applied after each FCL layer to regularize the model and avoid overfitting.

The performance of CNN models is highly influenced by their hyperparameters, such as the number of convolutional and fully connected layers, filter sizes, learning rate, etc. Careful selection of these parameters is essential to achieve optimal results. In ML, this process-known as hyperparameter fine-tuning-typically involves testing various combinations to identify the most effective configuration. To achieve optimal ML models and ensure a fair comparison with the statistical approach, hyperparameters are fine-tuned using Ray Tune \cite{liaw2018tune} on the validation dataset described in Section \ref{sec:data}. For the mono-label approach, since $N$ independent models are trained for $N$ radionuclides in each scenario, fine-tuning all models would be computationally intensive. Therefore, the same hyperparameters optimized for the multi-label model were used, and only the models with significantly worse performance compared to the multi-label model were further fine-tuned. Additional details on training the ML models are provided on the \href{https://github.com/triem1998/BenchmarkGamma/}{GammaBench} GitHub repository.

The performance of the ML models for the identification task highly depends on the threshold value that applies to the output of the neural network. Typically, this threshold is set to 0.5 \cite{liang2019rapid,daniel2020automatic,kim2019multi,turner2021convolutional,kim2025deep}, and when the output is greater than this value, the radionuclides are predicted to be present in the radioactive source. For a fair comparison with the statistical approach discussed in the next section, which depends on an expected false positive rate, this has been adjusted to achieve a false positive rate close to the desired value by applying the model to the validation set. In the following experiments, the expected FPR used in this work is set to 1$\%$.

For the quantification task, the same CNN architecture is used, with the output representing the mixing weights of all radionuclides. The loss function is the BCE with the softmax activation. As in the identification task, hyperparameters are optimized using Ray Tune.

\subsection{Spectral unmixing method}

\subsubsection{Quantification}

In $\gamma$-ray analysis, the spectral unmixing method aims to decompose an observed spectrum $y$ into individual spectral signatures of various radionuclides and Bkg. Mathematically, an observed spectrum $y$ is modeled by a Poisson distribution $y \sim \mathcal{P}(Xa)$, where, $X=[X_{Bkg},X_2,..,X_N] \in \mathbb{R}^{M\times N}$ stands for the matrix of spectral signatures (including Bkg), and $a \in \mathbb{R} ^{N}$ is the vector of counting. In this work, Bkg is supposed to be known. 

The cost function can be defined as the negative log-likelihood of the Poisson mixture model described in Eq.\ref{eq:poisson}:
\begin{equation}
L(y,X,a)=\sum_{m=1}^M ((Xa)_m - y_m log ((Xa)_m))
\label{eq:divergence}
\end{equation}
When the spectral signatures $X$ are known, the spectral unmixing problem involves estimating the non-negative vector $a$ that minimizes the cost function $L$:
\begin{equation} 
         \hat{a} =\argmin_{a\geq 0}  L(y,X,a) 
\label{eq:pb1}
\end{equation}
To efficiently solve this problem, the non-negative Poisson unmixing (NNPU) algorithm \cite{andre2021metrological} was developed, utilizing the multiplicative update rule.

However, in many practical scenarios, the spectral signatures $X$ are unknown and can be modified depending on the measurement conditions. For instance, in the reference \cite{phan2024hybrid}, spectral signatures are mostly deformed due to attenuation and Compton scattering when the radioactive source is placed in a steel sphere of varying thickness. Addressing such spectral variability requires building a surrogate model of the spectral signatures that captures their variability. To this end, a particular ML model called Interpolating AutoEncoder (IAE) \cite{bobin2023autoencoder, phan2024hybrid}) can be employed to model the spectral deformation, which can be derived from Geant4 simulations. The advantage of this IAE model is that it can be used in a generative way; the spectral signatures $X_{j \ge 2}$ can be modeled by a non-linear function of a latent variable $\lambda$ learned by IAE: $X_{j \ge 2} \approx f(\lambda) \, ; \lambda \in [0,1] $. Mathematically, the spectral unmixing problem is reformulated to jointly estimate the latent variable $\lambda$ (capturing spectral deformation) and the counting vector $a$ that minimizes the cost function $L$ :
\begin{equation} 
         \hat{\lambda},\hat{a} =\argmin_{\lambda \in [0,1],a\geq 0}  L(y, f(\lambda),a)  \quad  ;  \quad  \hat{X}_{j \ge 2}= f (\hat{\lambda} ) 
\label{eq:pb2}
\end{equation}
The semi-blind spectral unmixing (SEMSUN) algorithm \cite{phan2024hybrid} (\href{https://github.com/triem1998/SemSun/}{Github}) was developed to solve this above optimization problem.

For the spectral shift, $X$ can be modeled by a non-linear function of a shift factor $\alpha$: $X \approx g(\alpha)$. Therefore, the spectral unmixing problem is as follows:
\begin{equation} 
         \hat{\alpha},\hat{a} =\argmin_{\alpha ,a\geq 0}  L(y, g(\alpha),a)  \quad  ;  \quad  \hat{X}= g (\hat{\alpha} ) 
\label{eq:pb3}
\end{equation}
This optimization problem is similar to the spectral variability case and can be solved using the same algorithm. Additional details are provided in \ref{sec:shift}. 

\subsubsection{Identification}
The quantification result can be directly used to identify the radionuclide by calculating the decision threshold based on radionuclide counting. However, minimizing the cost function in Eq.\ref{eq:divergence} for the quantification problem often leads to noisy solutions. Indeed, additional radionuclides in the library, which are actually not present in the radioactive source, will tend to make the algorithm overfit the noise of the observed spectrum. This overfitting can lead to an under-estimation of the counting for all radionuclides and then lead to some false positive (alarm) detections. 

In order to overcome these issues, extra regularisation is necessary. A common approach involves enforcing sparsity on the counting vector $a$. Penalization of the number of identified radionuclides is equivalently reformulated as penalizing the non-zero entries of the counting vector $a$. To that end, the optimization problem can be reformulated as:
\begin{equation} 
\hat{\theta} =\argmin_{\theta \in C}  L(y,\theta) \text{  subject to } ||a||_0 = K
\label{eq:pb_iden}
\end{equation}
where $\theta$ is $(a)$ for known spectral signatures, $\theta$ is $(\lambda,a)$ for spectral deformation or $\theta$ is $(\alpha,a)$ for spectral shift. The constraint space $C$ includes all feasible parameter values $\theta$, satisfying the conditions outlined in the previous sections.

To solve this problem, algorithms such as Poisson Orthogonal Matching Pursuit (P-OMP) \cite{andre2021metrological} and Model selection with spectral variability based on manifold learning (MoSeVa)\cite{phan2025automatic} have been developed for cases involving known or deformed spectral signatures. These algorithms implicitly enforce sparsity thanks to a greedy model selection procedure, where models with increasing dimensionality are tested, and a statistics-based criterion is employed to stop the selection procedure. By using the likelihood ratio test (LRT) \cite{wilks1938large}, this approach can correctly identify the radionuclide with the ability to control the false positive rate (see \cite{andre2021metrological} for more details). For the case of spectral shift, the problem formulation is similar to that of spectral deformation, allowing the same type of algorithm to be employed.

Note that in Scenarios 2 and 3, errors in modeling spectral signatures can lead to false positives at high statistics, although the corresponding radionuclide counting is very low. To keep the false positive rate close to the expected value, an additional constraint is introduced: a radionuclide is considered present only if it contributes at least 1$\%$ of the total counts $\sum_{i=2}^N a_i$ from all radionuclides, excluding background. This threshold is based on prior knowledge used during dataset generation.

\section{Results}
\subsection{Identification performances}\label{sec:iden}
\subsubsection{Scenario 1: Known spectral signatures }
Table \ref{table:false_pos_exp1} shows the identification performance of the multi-label and mono-label CNN techniques, compared with the spectral unmixing method. The FPR obtained by all methods is close to the expected value of 1$\%$. Concerning the other metrics, the spectral unmixing method gives a lower FNR, as well as higher accuracy and PPR than the ML techniques. As a result, the unmixing method outperforms the ML methods in terms of identification performance.

\begin{figure}[H]
    \centering
      \includegraphics[scale=0.18]{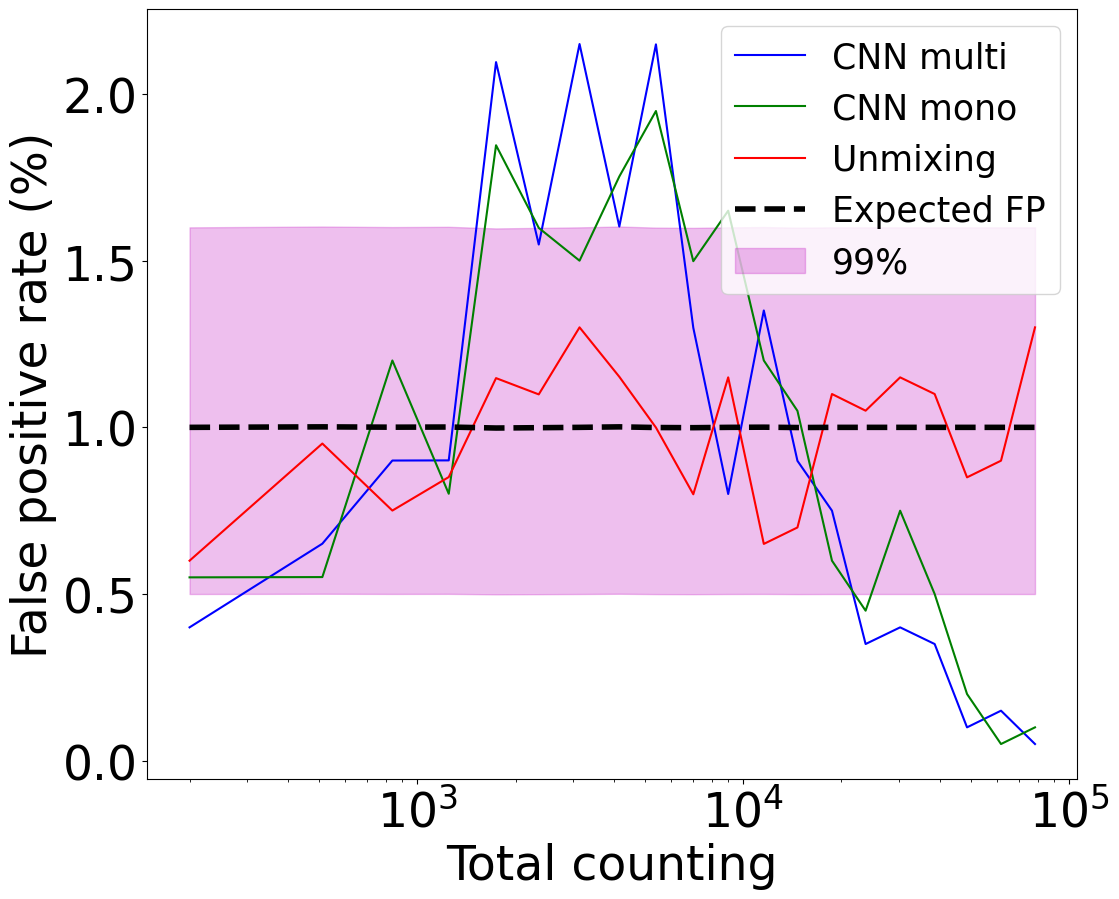}
    \includegraphics[scale=0.18]{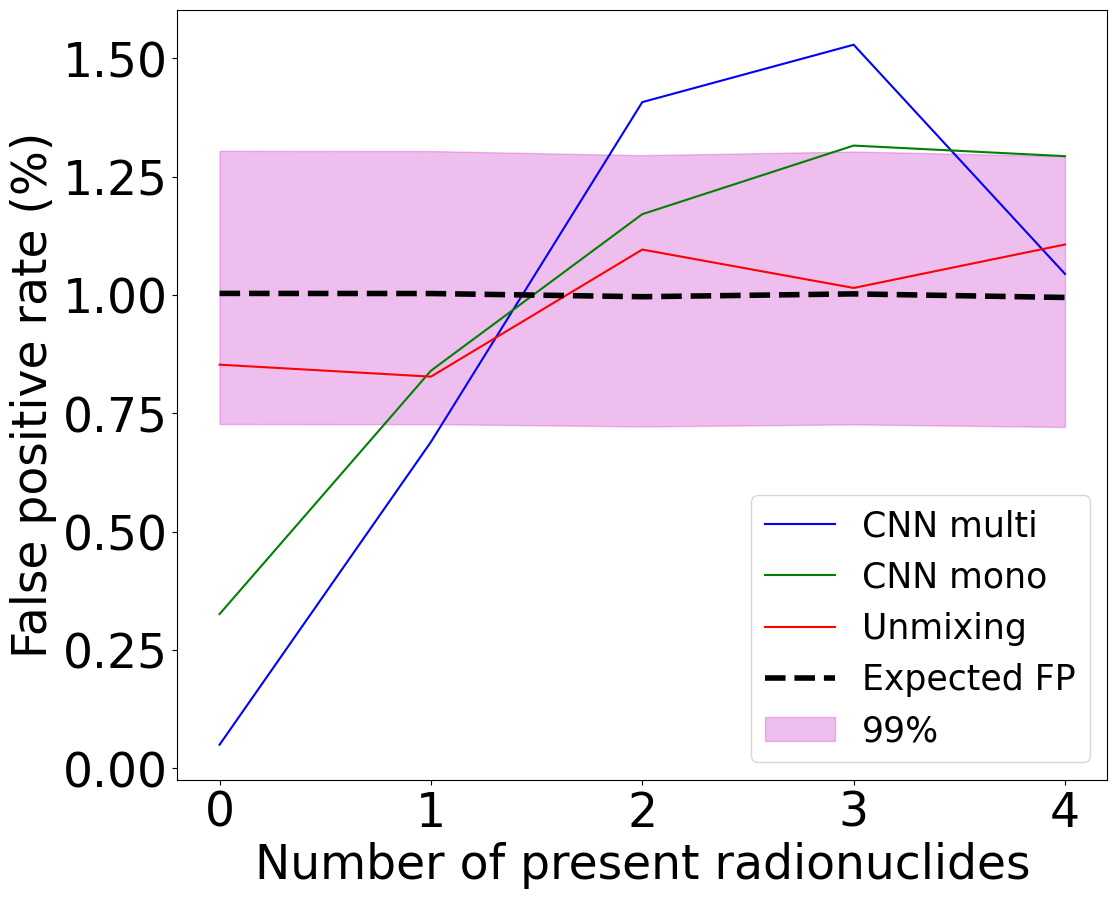}
    \caption{Scenario 1: FPR as a function of total counting (left) and number of present radionuclides (right).}
      \label{figure:fpr_exp1}
\end{figure}

\autoref{figure:fpr_exp1} illustrates the FPR of all methods as a function of total counting and the number of present radionuclides. While the average FPR of the ML methods remains close to the expected level across the entire dataset, differences from the expected value appear when the total counting or the number of present radionuclides varies. In contrast, the spectral unmixing method, based on a statistical hypothesis test, maintains the FPR within the error bars of the expected value. 

\begin{figure}[H]
    \centering
      \includegraphics[scale=0.3]{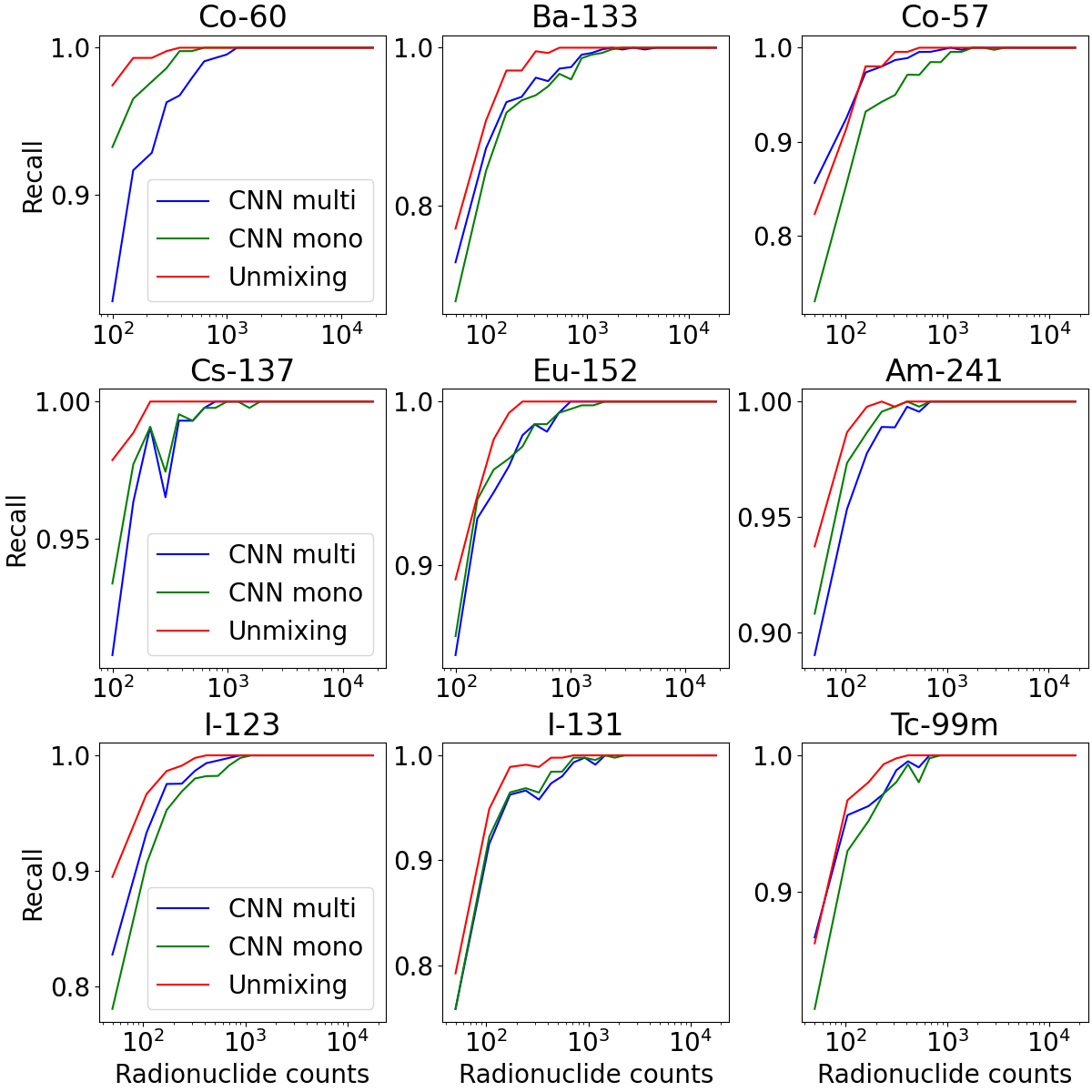}
    \caption{Scenario 1: Recall (1-FNR) for each radionuclide as a function of radionuclide counts.}
      \label{figure:fnr_exp1}
\end{figure}

To provide a clearer understanding of the FNR, \autoref{figure:fnr_exp1} shows the recall (1-FNR) of each radionuclide as a function of its counting. When the statistics are sufficient (counting greater than 1000 per radionuclide), both methods can accurately identify radionuclides. On the other hand, at lower statistical levels, the spectral unmixing method has a better detection capability, as it is based on physical properties such as the Poisson distribution and the linear nature of the mixing model. 

\begin{table}[H]
\centering
\small
\begin{tabular}{lrrrrr}
\toprule
{} &  Accuracy &    PPR &  FPrR &  FPR &  FNR \\
\midrule
CNN multi &    99.49 &  95.53 &  4.47 &  0.94 &  3.60 \\
CNN mono  &    99.44 &  95.11 &  4.89 &  0.99 &  3.96 \\
Unmixing  &    99.68 &  97.19 &  2.81 &  0.98 &  1.88 \\
\bottomrule
\end{tabular}
\caption{Identification performance for the test dataset for Scenario 1.  }
\label{table:false_pos_exp1}
\end{table}

\subsubsection{Scenario 2: Spectral deformation}
The first numerical experiment examines the robustness of these methods in the presence of spectral variability. Specifically, the spectral signatures employed are the known (incorrect) signatures from Scenario 1 tested on a dataset including spectral deformation. Pre-trained CNN models based on the known spectral signatures from Scenario 1 and the P-OMP algorithm are employed for evaluation. The test dataset for this case introduces spectral variability by varying the steel thickness in a small range, from 0.001 mm to 1 mm. Even in this range, spectral signatures are considerably deformed, particularly for low-energy peaks that are absorbed by the attenuation phenomenon. Table \ref{table:false_pos_exp2_test} shows the result of both methods in this case. Since the spectral signatures used are incorrect and do not take into account the spectral deformation, the quality of the results for all methods decreases significantly. The spectral unmixing method is more sensitive than ML methods since it tries to fit the observed $\gamma$-spectrum using the individual spectral signatures. When the spectral signatures are incorrect, the spectral unmixing algorithm tends to fit the spectrum by the spectral signatures of non-present radionuclides, leading to a significantly high FPR. In contrast, ML techniques are similar to the region of interest, highly depending on the peak's position, making them less sensitive to incorrect spectral signatures.

To ensure more robust identification, spectral variabilities must be taken into account. For this purpose, new CNN models are trained using the dataset described in Section \ref{sec:data}, which includes spectral variability. To that end, the MoSeVa algorithm is employed instead of P-OMP to tackle the spectral unmixing task. Table \ref{table:false_pos_exp2} features the identification evaluation of these methods. Similar to Scenario 1, all methods achieve the FPR close to the expected values. Additionally, the spectral unmixing method achieves a lower FNR and higher accuracy and PPR than ML methods.

\begin{figure}[H]
    \centering
      \includegraphics[scale=0.18]{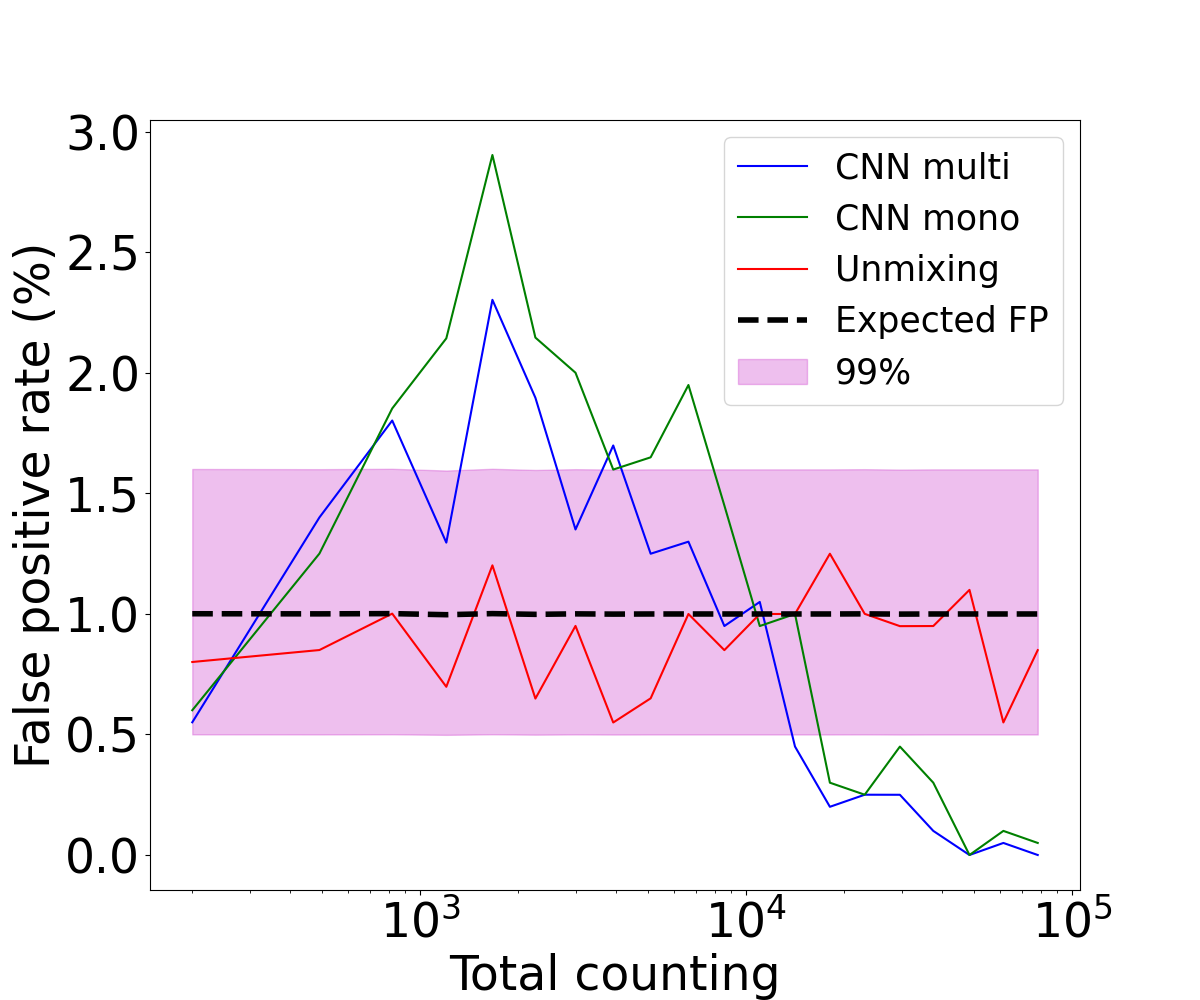}
    \includegraphics[scale=0.18]{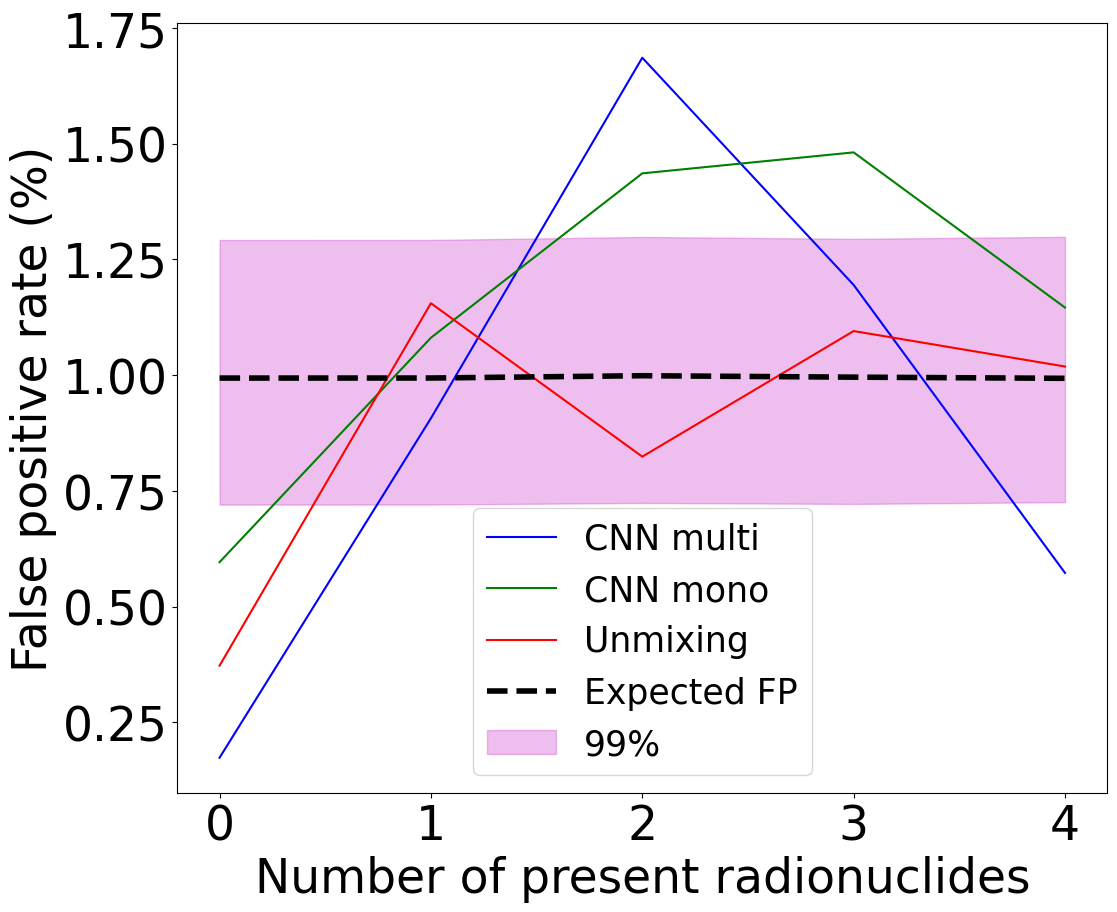}
    \caption{Scenario 2: FPR as a function of total counting (left) and number of present radionuclides (right).}
      \label{figure:fpr_exp2}
\end{figure}

\begin{figure}[H]
    \centering
      \includegraphics[scale=0.3]{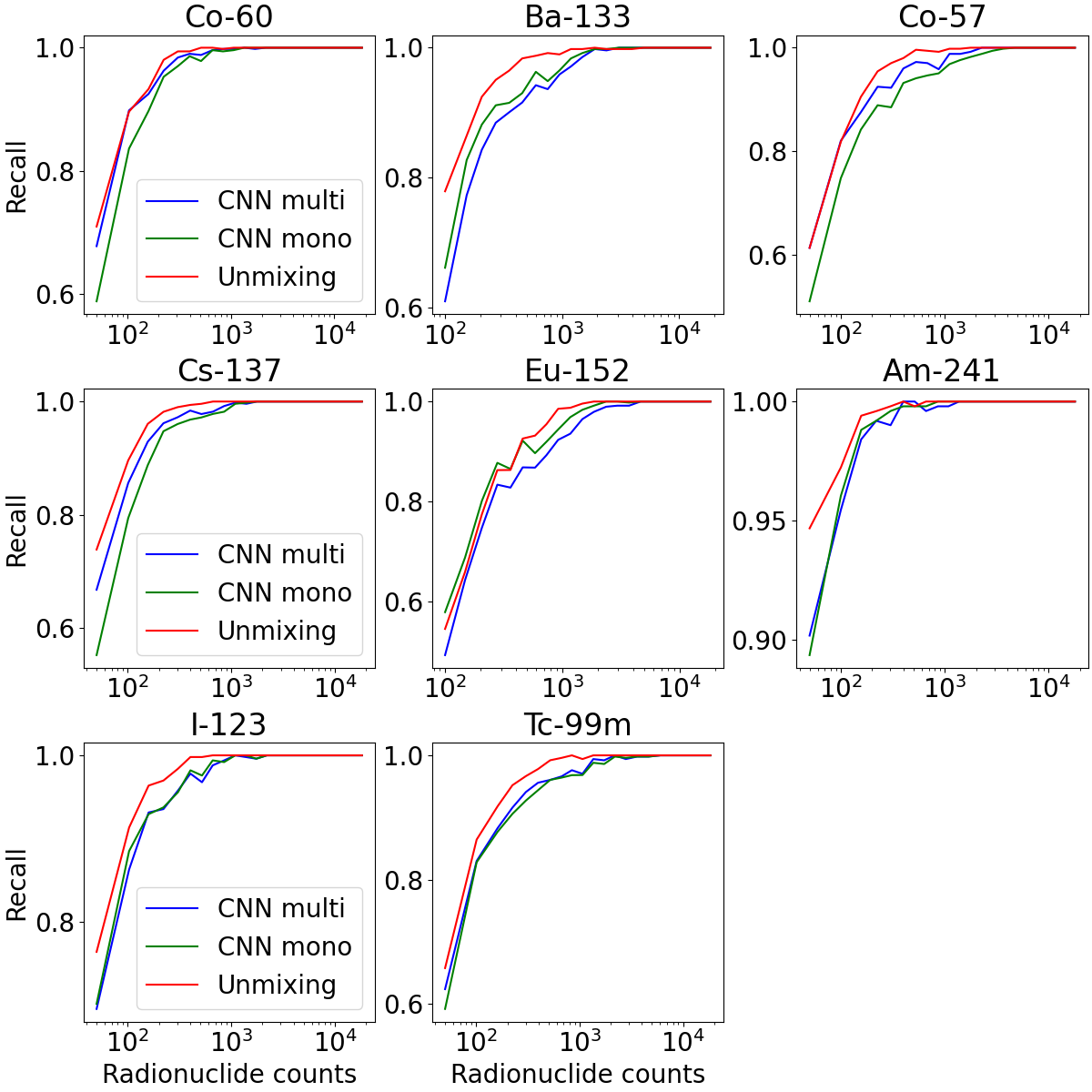}
    \caption{Scenario 2: Recall (1-FNR) for each radionuclide as a function of radionuclide counts.}
      \label{figure:fnr_exp2}
\end{figure}

\autoref{figure:fpr_exp2} illustrates the FPR of all methods as a function of total counting and the number of present radionuclides. When these parameters vary, the FPR obtained by ML methods occasionally exceeds the expected values. Conversely, the spectral unmixing method keeps the FPR below the expected value. \autoref{figure:fnr_exp2} displays the recall of each radionuclide as a function of its counting. With sufficient statistics, both methods accurately identify radionuclides. At lower statistical levels, the spectral unmixing method demonstrates superior detection capability.

\begin{figure}[H]
    \centering
      \includegraphics[scale=0.3]{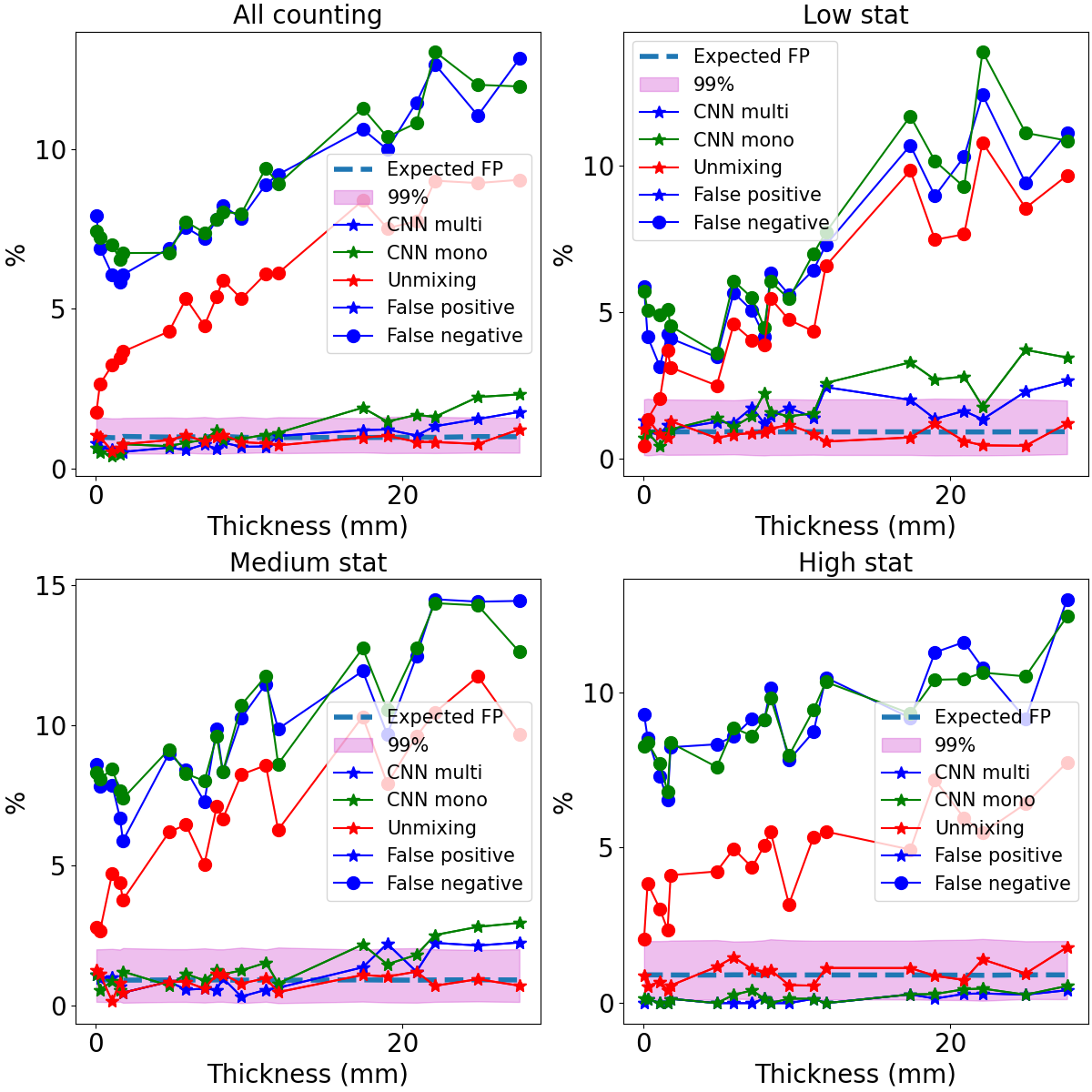}
    \caption{Scenario 2: FPR and FNR for different thickness values at various counting levels. The top-left panel shows results for all counting levels combined. The other panels show results for low, medium, and high statistics, respectively.}
      \label{figure:thick_exp2}
\end{figure}

Identification performance for all thickness values and different counting levels are shown in \autoref{figure:thick_exp2}. As the thickness increases, attenuation effects become more important, leading to an increase in FNR for all methods. With regard to the total counting, the FPR of the ML methods is higher at low statistical levels but decreases significantly when counts increase. Notably, high steel thickness values (greater than 20 mm) tend to be the most difficult scenarios for radionuclide identification for all methods.

\begin{table}[H]
\centering
\small
\begin{tabular}{lrrrrr}
\toprule
{} &  Accuracy &    PPR &  FPrR &   FPR &   FNR \\
\midrule
CNN multi &    97.94 &  85.04 &  14.96 &   4.53 &  11.02 \\
CNN mono  &    98.31 &  87.55 &  12.45 &   3.62 &   9.27 \\
Unmixing  &    97.20 &  82.82 &  17.18 &  12.22 &   5.86 \\

\bottomrule
\end{tabular}
\caption{Scenario 2: Identification performance in the case of spectral variability with known incorrect spectral signatures. Steel thicknesses range from 0.001 mm to 1 mm. }
\label{table:false_pos_exp2_test}
\end{table}

\begin{table}[H]
\centering
\small
\begin{tabular}{lrrrrr}
\toprule
{} &  Accuracy &    PPR &  FPrR &   FPR &   FNR \\
\midrule
CNN multi &    98.74 &  90.68 &              9.32 &  0.92 &  8.62 \\
CNN mono  &    98.66 &  90.15 &              9.85 &  1.15 &  8.86 \\
Unmixing  &    99.12 &  93.55 &              6.45 &  0.89 &  5.70 \\
\bottomrule
\end{tabular}
\caption{Scenario 2: Identification performance in the case of spectral variability with unknown spectral signatures. Steel thickness values range from 0.001 mm to 30 mm. }
\label{table:false_pos_exp2}
\end{table}

\subsubsection{Scenario 3: Gain shift}
The robustness of ML and spectral unmixing methods is evaluated with respect to gain shift. Specifically, the spectral signatures employed are the known (incorrect) signatures from Scenario 1 tested on a dataset that includes gain shifts. Pre-trained CNN models based on the known spectral signatures from Scenario 1 and the P-OMP algorithm are used for evaluation. The test dataset includes the spectral shift with the shift factor varying from -2 to 2$\%$. Table \ref{table:false_pos_exp3_test} shows the results of both methods in this case. As the spectral signatures used are incorrect and do not take into account the spectral shift, all methods show a significant decrease in performance. The spectral unmixing method is more sensitive than the ML methods, similar to the behavior observed in Scenario 2.

\begin{figure}[H]
    \centering
      \includegraphics[scale=0.18]{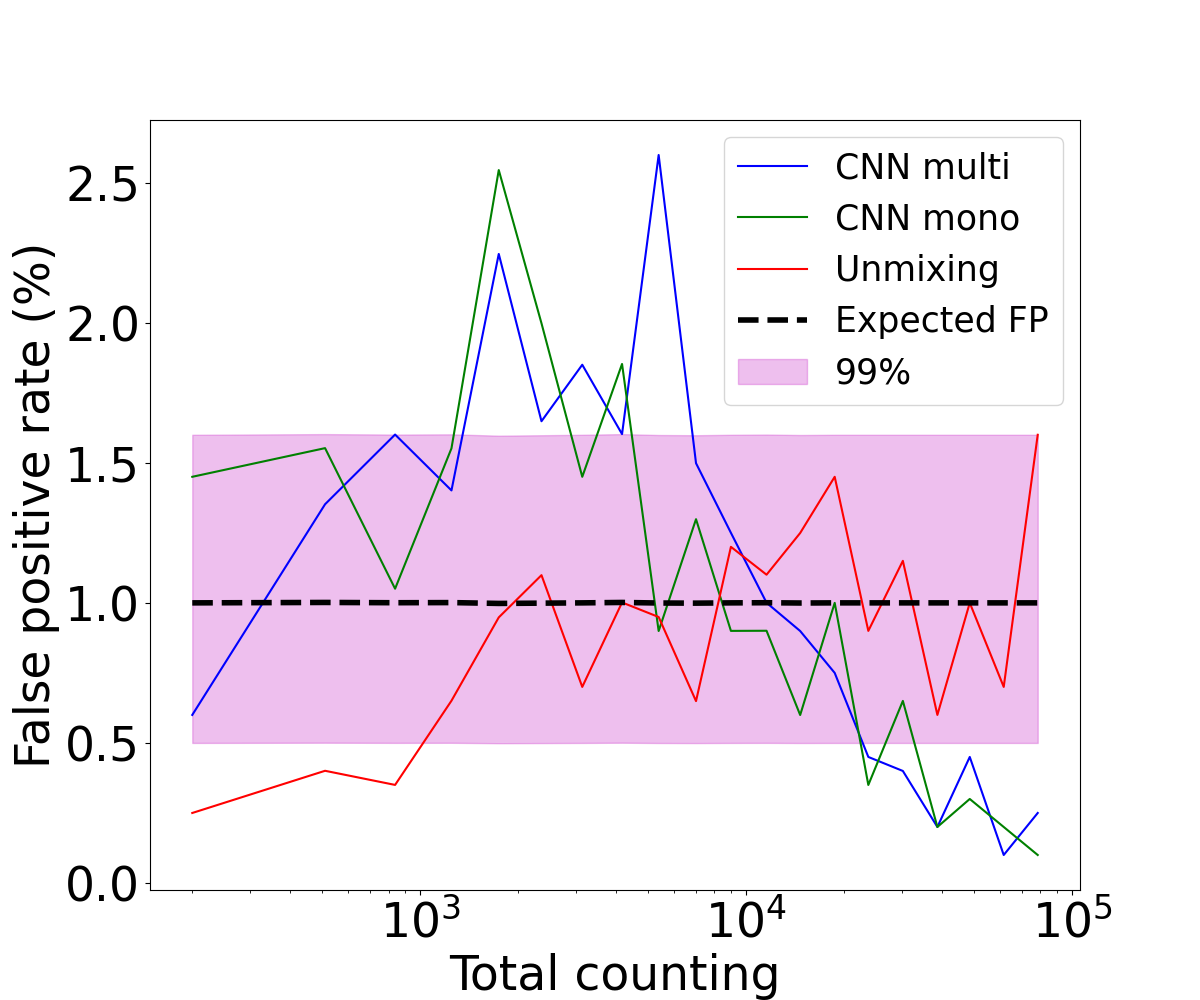}
    \includegraphics[scale=0.18]{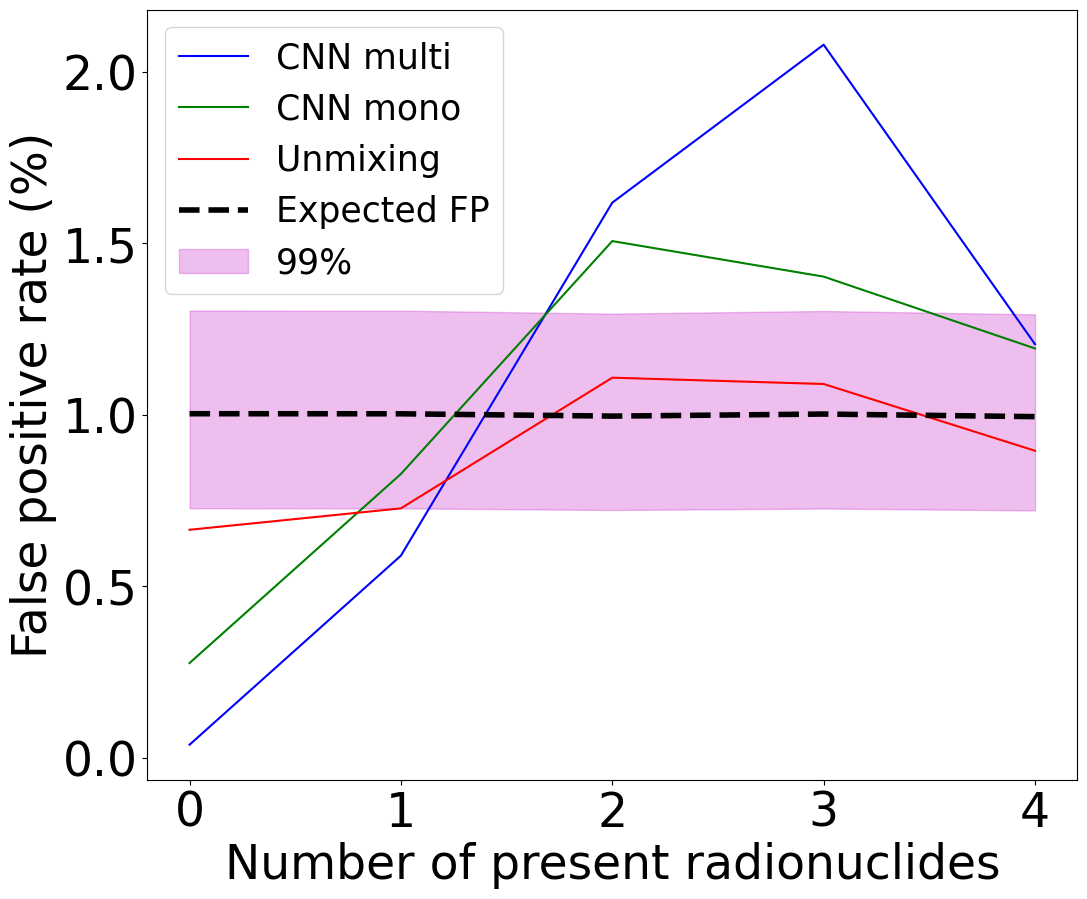}
    \caption{Scenario 3: FPR as a function of total counting (left) and number of present radionuclides (right).}
      \label{figure:fpr_exp3}
\end{figure}

\begin{figure}[H]
    \centering
      \includegraphics[scale=0.25]{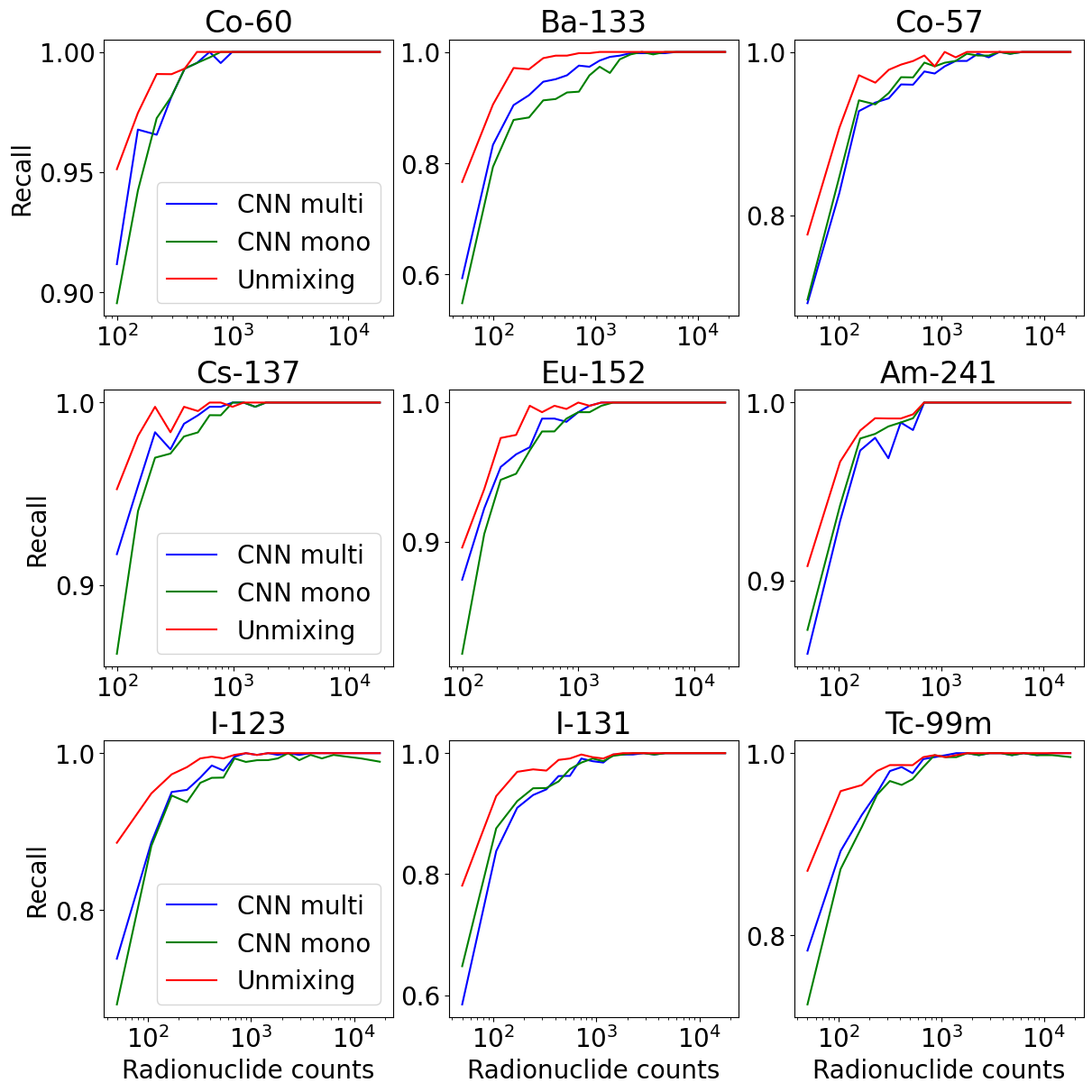}
    \caption{Scenario 3: Recall (1-FNR) as a function of radionuclide counts.}
      \label{figure:fnr_exp3}
\end{figure}

To ensure robust identification performance, the spectral shift must be considered. To this end, new CNN models are trained using the dataset described in Section 2.1, which includes gain shifts. For the spectral unmixing method, the P-OMP algorithm with gain shift consideration described in \ref{sec:shift} is employed, replacing the original P-OMP. Table \ref{table:false_pos_exp3} summarizes the identification results of these methods. Similarly to Scenario 1, all methods achieve an FPR close to the expected value, and the spectral unmixing method achieves a lower FNR and higher accuracy and PPR than ML methods. \autoref{figure:fpr_exp3} demonstrates that the spectral unmixing method allows the FPR to be lower than the expected value while the FPR obtained by ML methods occasionally exceeds the expected values. \autoref{figure:fnr_exp3} shows the recall of each radionuclide as a function of its counting. At lower statistical levels, the spectral unmixing method demonstrates superior detection capability.

\begin{figure}[H]
    \centering
      \includegraphics[scale=0.25]{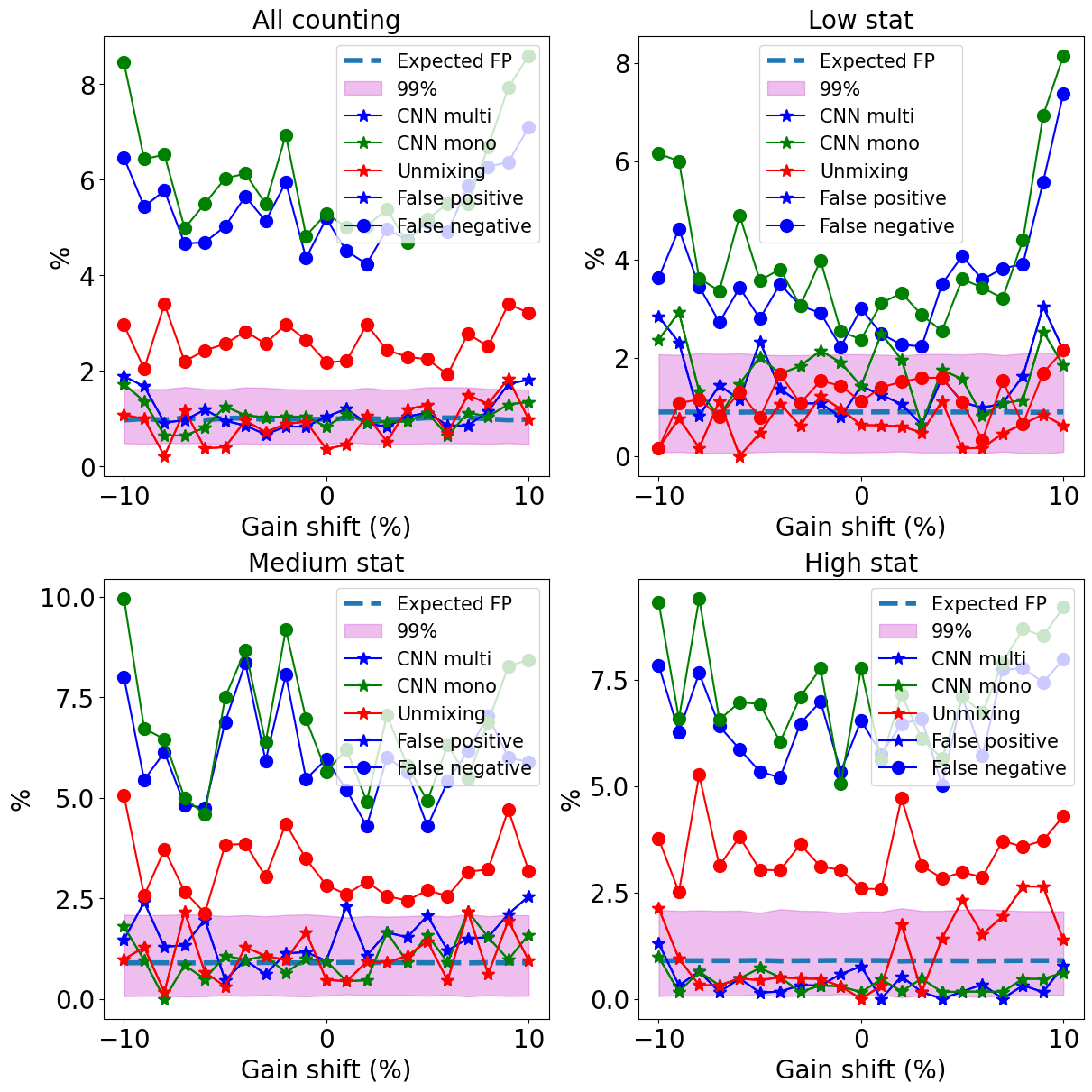}
    \caption{Scenario 3: FPR and FNR for different gain shift factors with different counting levels. The top-left panel shows results for all counting levels combined. The other panels show results for low, medium, and high statistics, respectively.}
      \label{figure:shift_exp3}
\end{figure}

Identification performance for different shift factors and counting levels are shown in \autoref{figure:shift_exp3}. High gain shift factors (close to 10$\%$) present the most challenging cases for radionuclide identification for ML methods, as the spectral signatures are significantly modified. 

\begin{table}[H]
\centering
\small
\begin{tabular}{lrrrrr}
\toprule
{} &  Accuracy &    PPR &  FPrR &   FPR &   FNR \\
\midrule
CNN multi &    98.21 &  85.52 &  14.48 &  10.64 &  4.46 \\
CNN mono  &    97.82 &  82.49 &  17.51 &  13.72 &  4.59 \\
Unmixing  &    95.15 &  67.40 &  32.60 &  31.10 &  2.22 \\

\bottomrule
\end{tabular}
\caption{Scenario 3: Identification performance in the case of gain shift with known incorrect spectral signatures. The shift factor is between -2$\%$ and 2$\%$.  }
\label{table:false_pos_exp3_test}
\end{table}

\begin{table}[H]
\centering
\small
\begin{tabular}{lrrrrr}
\toprule
{} &  Accuracy &    PPR &  FPrR &   FPR &   FNR \\
\midrule
CNN multi &    99.27 &  93.75 &  6.25 &  1.11 &  5.35 \\
CNN mono  &    99.19 &  93.13 &  6.87 &  1.04 &  6.00 \\
Unmixing  &    99.60 &  96.59 &  3.41 &  0.90 &  2.60 \\

\bottomrule
\end{tabular}
\caption{Scenario 3: Identification performance in the case of gain shift with unknown spectral signatures. The shift factor is between -10$\%$ and 10$\%$.   }
\label{table:false_pos_exp3}
\end{table}

\subsection{Quantification performance}
\begin{figure}[H]
    \centering
      \includegraphics[scale=0.25]{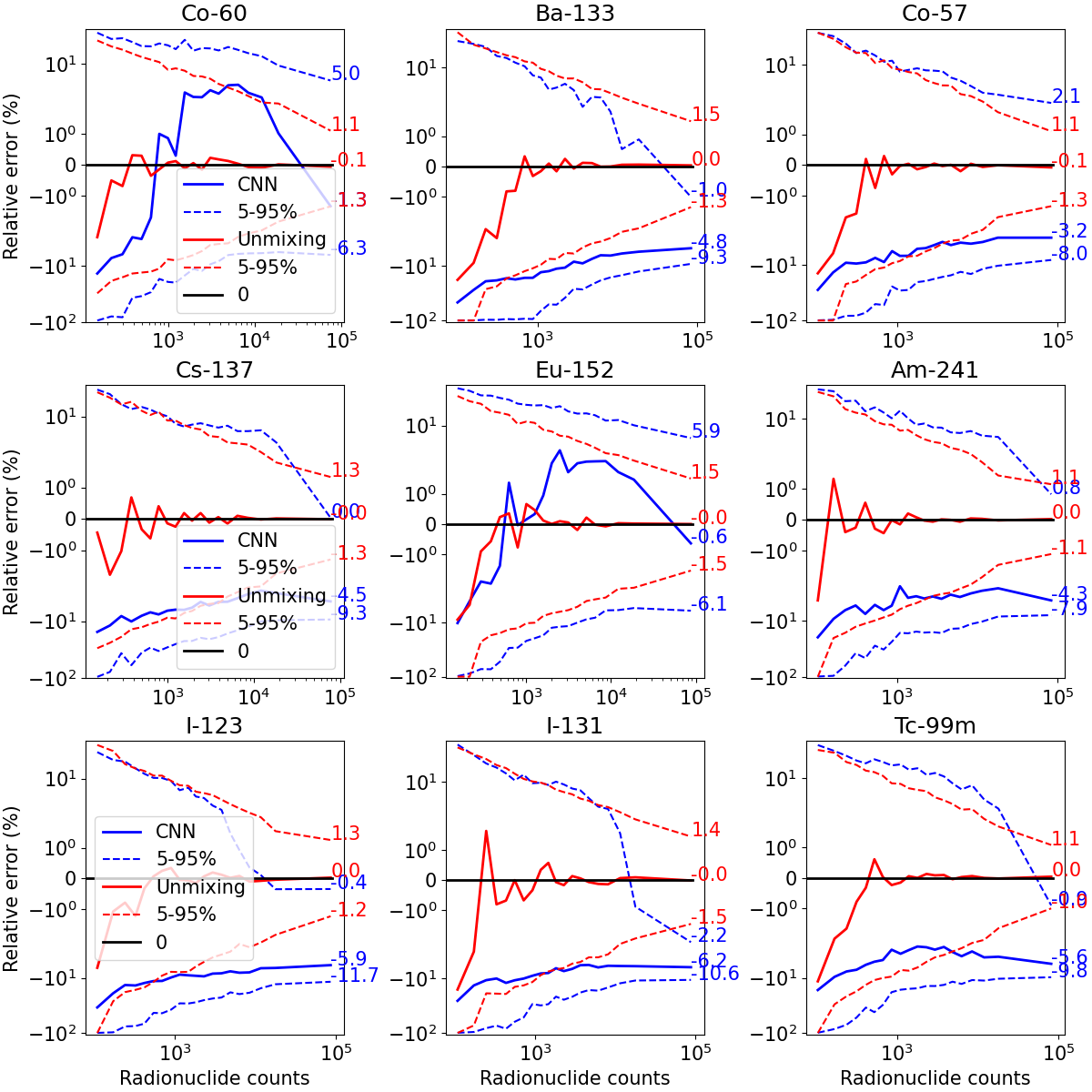}
    \caption{Scenario 1: Relative error of the estimated counting for each radionuclide. For each radionuclide, the 5-95$\%$ range represents the interval between the 5th and 95th percentiles of the relative errors, with the continuous line indicating the average value.    }
      \label{figure:rel_exp1}
\end{figure}

\begin{figure}[H]
    \centering
      \includegraphics[scale=0.25]{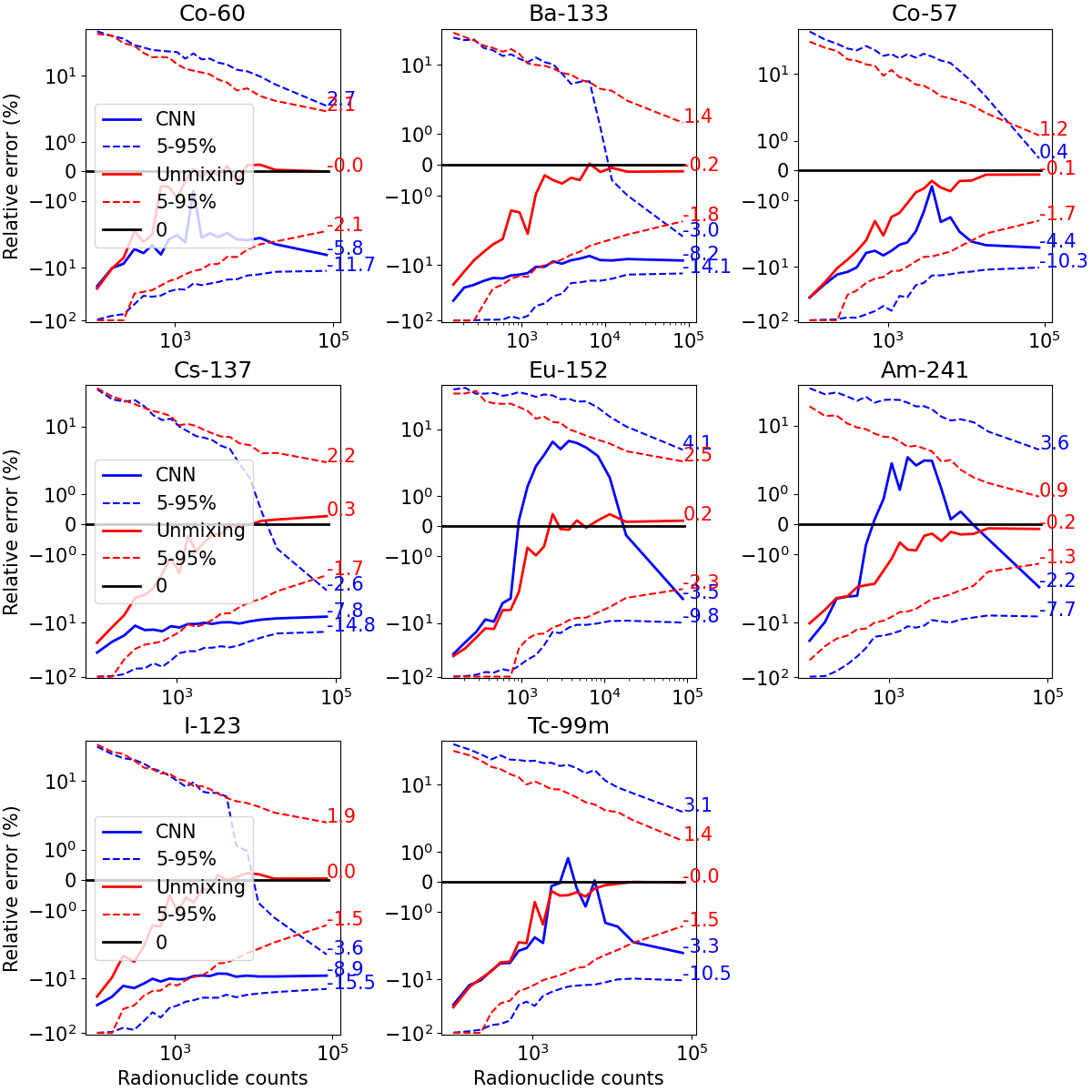}
    \caption{Scenario 2: Relative error of estimated counting for each radionuclide.}
      \label{figure:rel_exp2}
\end{figure}

\begin{figure}[H]
    \centering
      \includegraphics[scale=0.25]{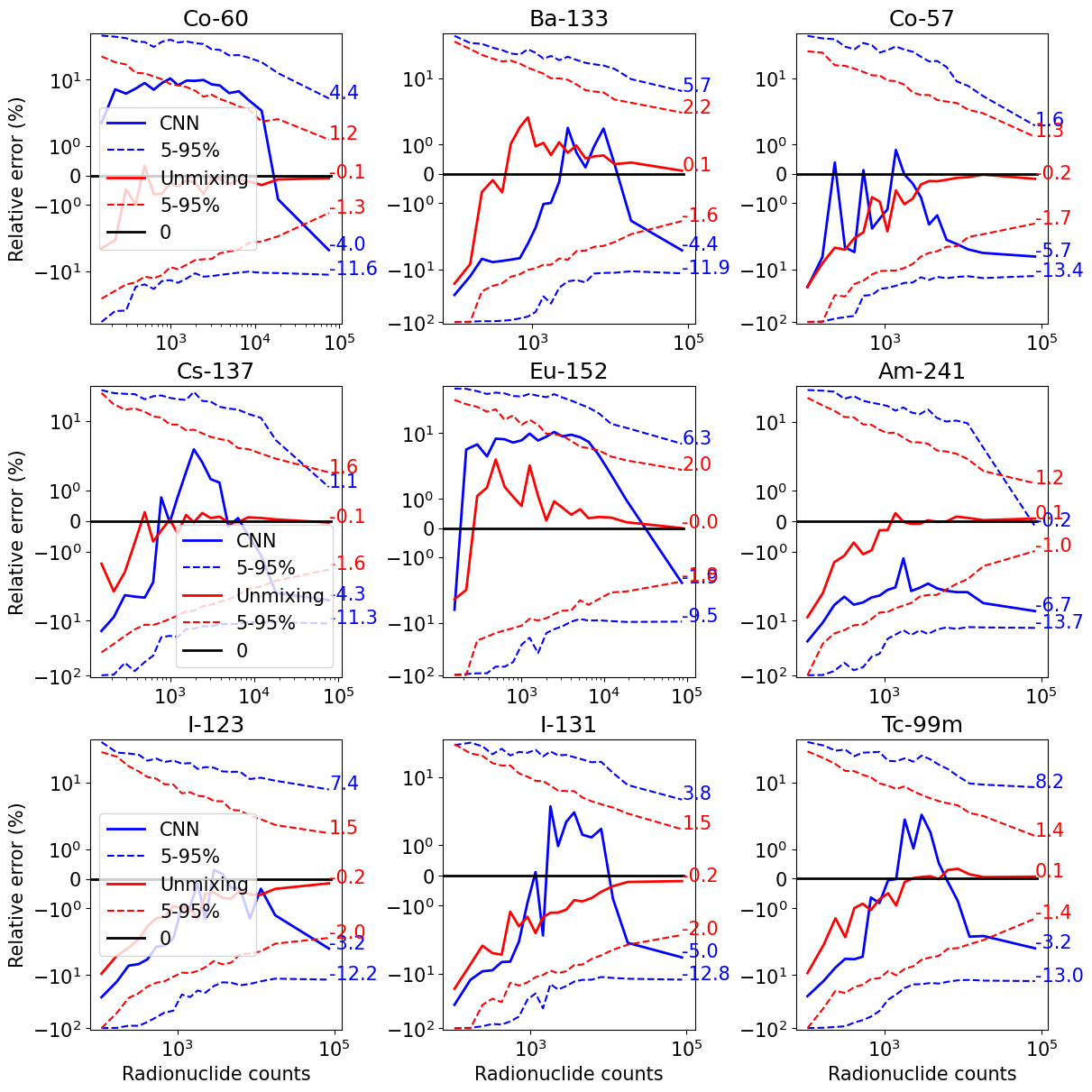}
    \caption{Scenario 3: Relative error of estimated counting for each radionuclide.}
      \label{figure:rel_exp3}
\end{figure}

Table \ref{table:quan} shows the quantification metrics MSE and RAE for both the ML and spectral unmixing methods across the three scenarios. The spectral unmixing method outperforms the ML method, achieving the lowest error for all metrics. Fig.\ref{figure:rel_exp1}, \ref{figure:rel_exp2} and \ref{figure:rel_exp3}  display the relative error of estimated counting for each radionuclide in the three scenarios. The spectral unmixing method consistently provides estimated radionuclides counting with means close to the expected values and very low variability. In contrast, the ML method exhibits significant bias and high variability in its estimates across all scenarios. For instance, for Scenario 1 illustrated in \autoref{figure:rel_exp1}, the mean of relative error of spectral unmixing method is close to 0 with 90$\%$ of relative error falling below 1.5$\%$ when the radionuclide counting is high ($10^5$) and below  10$\%$ when the radionuclide counting is greater than 1000. For the ML method, the estimated counting is less accurate, with the mean of relative error being 6$\%$ and 90$\%$ of relative error falling below 11$\%$ even at radionuclide counting of $10^5$. 

Similar to the robustness evaluation in Section \ref{sec:iden}, the quantification performance is also calculated using the known incorrect spectral signatures, as shown in Table \ref{table:quan_test}. As demonstrated earlier, the spectral unmixing method is more affected than the ML method for the identification task when the spectral signatures are incorrectly modeled. However, for the quantification task, the spectral unmixing method outperforms the ML method, yielding lower errors for all metrics and scenarios. 

\begin{table}[H]
\centering
\small
\begin{tabular}{lrrrrrr}
\toprule
{} & \multicolumn{2}{c}{Scenario 1}  & \multicolumn{2}{c}{Scenario 2} & \multicolumn{2}{c}{Scenario 3} \\ 
\midrule
{} &  MSE(1e-5) &    RAE($\%$) &  MSE(1e-5)&   RAE($\%$) &   MSE(1e-5) & RAE($\%$) \\
\midrule
CNN  &    14.1 &  10.24 &              29.5 &  13.21 &  20.1 & 12.27 \\
Unmixing  &  3.3   &  4.42 &              11.8 &  7.46 &  4.9 &5.03 \\
\bottomrule
\end{tabular}
\caption{Quantification performances.    }
\label{table:quan}
\end{table}

\begin{table}[H]
\centering
\small
\begin{tabular}{lrrrr}
\toprule
{}  & \multicolumn{2}{c}{Scenario 2} & \multicolumn{2}{c}{Scenario 3} \\ 
\midrule
{} &   MSE(1e-5)&   RAE($\%$) &   MSE(1e-5) & RAE($\%$) \\
\midrule
CNN  &             163.2 &  21.11 &  28.4 & 13.65 \\
Unmixing   &              82.7 &  14.61 &  7.8 &6.89 \\
\bottomrule
\end{tabular}
\caption{Quantification performance used known incorrect spectral signatures.    }
\label{table:quan_test}
\end{table}

\section{Conclusion}
This work introduces an open-source benchmark with data, code and metrics to compare the automatic identification and quantification performance of the end-to-end ML and statistical unmixing methods in $\gamma$-ray spectrometry. These methods are evaluated across three scenarios: (1) spectral signatures are assumed to be known, (2) spectral signatures are deformed due to physical phenomena like Compton scattering and attenuation, and (3) gain shift. A library of Geant4 simulated spectral signatures for nine radionuclides, alongside an experimental natural background, is utilized in this study. 

Regarding identification performance, the statistical unmixing approach consistently outperforms the ML approaches across all three scenarios in terms of accuracy and false prediction rate, especially at low statistics. Additionally, the statistical method effectively controls the false alarm rate, keeping it close to the predefined value, which is less robust for ML methods. However, the performance of the statistical approach can be impacted when spectral signatures are not modeled correctly. 

For quantification, the statistical approach provides accurate estimates of radionuclide counting or mixing weights, while the ML methods deliver less satisfactory results. When measurement conditions are not well-defined or difficult to model, the statistical approach still yields better outcomes than ML.

In conclusion, the statistical unmixing approach is most effective for the identification task when spectral signatures are known or spectral variability is correctly modeled. In cases where measurement conditions are not well-defined or challenging to model, end-to-end ML provides a suitable alternative. For quantification tasks, the statistical method is the preferred choice.

\appendix

\section{Algorithm for gain shift}\label{sec:shift}
Let $s^{(s)}$ denote the shifted spectral signature of a radionuclide corresponding to a gain shift factor $\alpha$, and $s^{(r)}$ represents the reference spectral signatures (without shift). In this work, the form of relation function that induces the spectral shift is assumed to be known $e^{(s)}= h(e^{(r)},\alpha)= e^{(r)} \times (1-\alpha)$; only the parameter $\alpha$ needs to be estimated. Since this relation applies to energy rather than channel, we can construct an 'artificial' list with associated energies from the reference spectral signature $s^{(r)}$.

\begin{itemize}
    \item $s^{(r)}_i$ is the value for channel $i$ of reference spectral signature, corresponding to the energy range $[2i+20,2i+2+20]$ (binning of 2 keV per channel and a low-energy cut-off of 20 keV)
    \item The normalized spectrum $s^{(r)}$ is multiplied by a large number ({\it e.g,} $10^6$) to bring the spectrum down to high statistics: $s^{(*)}=s^{(r)} \times 10^6$. In this case, $s^{(*)}_i$ can be interpreted as the number of times the energy falls within the interval $[2i+20,2i+20+2]$.
    \item For each channel $i$, an energy list is defined as $\{2i+20,2i+20+d, 2i+20+2d,..,2i+20+2\}$, where the number of points corresponds to $s^{(*)}_i$.
    \item An energy list $\{\hat e^{(r)}_j\}$ can be obtained by cumulating all channels.
\end{itemize}

Using this energy list derived from the reference spectral signature, the shift function can be applied to compute the shifted energy list $\{ e^{(s)}_j\}$: $ e ^{(s)}_j=h(\hat e^{(r)}_j,\alpha)$. The spectral signature is then obtained by constructing a histogram from this list. Consequently, the shifted spectral signatures of all radionuclides $X^{(s)}$ can be modeled as a function of $\alpha$: $X^{(s)}=g(X^{(r)},\alpha)$.

This optimization problem is similar to the case of spectral deformation. To address this, the BCD minimization scheme implemented in SEMSUN is used. It is important to note that this problem is non-differentiable because it involves converting the energy list into a histogram (bin = $\sum 1_{2i<e_j<2i+2}$). The Nelder-Mead algorithm \cite{lagarias1998convergence}, which relies on direct search methods, is well-suited to solving this without requiring derivatives.

\newpage
 \bibliographystyle{elsarticle-num} 
 \bibliography{biblio}

\end{document}